\documentclass[runningheads]{llncs}
\usepackage[T1]{fontenc}
\usepackage{subcaption}
\usepackage{mathtools}
\usepackage{xcolor}
\newcommand\independent{\protect\mathpalette{\protect\independenT}{\perp}}
\def\independenT#1#2{\mathrel{\rlap{$#1#2$}\mkern2mu{#1#2}}}

\usepackage{mwe}
\usepackage[utf8]{inputenc}
\usepackage{amssymb}
\usepackage{graphicx}
\usepackage{mwe}
\usepackage{yhmath}
\usepackage{authblk}
\newcommand{\blue}{\textcolor[HTML]{1f77b4}}
\newcommand{\orange}{\textcolor[HTML]{ff7f0e}}
\newcommand{\green}{\textcolor[HTML]{2ca02c}}
\newcommand{\red}{\textcolor[HTML]{d62728}}
\newcommand{\purple}{\textcolor[HTML]{9467bd}}
\newcommand{\pink}{\textcolor[HTML]{e377c2}}
\newcommand{\cyan}{\textcolor[HTML]{17becf}}
\usepackage[symbol]{footmisc}

\title{Improving Image-Based Precision Medicine with Uncertainty-Aware Causal Models}

\author{Joshua Durso-Finley$^{1,4}$, Jean-Pierre Falet$^{1,3,4}$, Raghav Mehta$^{1,4}$, Douglas L. Arnold$^{1,3}$, Nick Pawlowski$^2$, Tal Arbel$^{1,4}$}
\authorrunning{Joshua Durso-Finley et al.}
\institute{$^1$Center for Intelligent Machines, McGill University. $^2$Microsoft Research. $^3$Montreal Neurological Institute, McGill University. $^4$MILA (Quebec AI institute).}
\begin{document}

\maketitle
\begin{abstract}
   Image-based precision medicine aims to personalize treatment decisions based on an individual's unique imaging features so as to improve their clinical outcome. Machine learning frameworks that integrate uncertainty estimation as part of their treatment recommendations would be safer and more reliable. However, little work has been done in adapting uncertainty estimation techniques and validation metrics for precision medicine. In this paper, we use Bayesian deep learning for estimating the posterior distribution over factual and counterfactual outcomes on several treatments. This allows for estimating the uncertainty for each treatment option and for the individual treatment effects (ITE) between any two treatments. We train and evaluate this model to predict future new and enlarging T2 lesion counts on a large, multi-center dataset of MR brain images of patients with multiple sclerosis, exposed to several treatments during randomized controlled trials. We evaluate the correlation of the uncertainty estimate with the factual error, and, given the lack of ground truth counterfactual outcomes, demonstrate how uncertainty for the ITE prediction relates to bounds on the ITE error. Lastly, we demonstrate how knowledge of uncertainty could modify clinical decision-making to improve individual patient and clinical trial outcomes.
   
\end{abstract}

\section{Introduction}

Precision medicine permits more informed treatment decisions to be made based on individual patient characteristics (e.g. age, sex), with the goal of improving patient outcomes. Deep causal models based on medical images can significantly improve personalization by learning individual, data-driven features to predict the effect of treatments.\footnote{See \cite{RueckertCausalReview} for a review on causality in medical imaging.} As a result, they could significantly improve patient outcomes, particularly in the context of chronic, heterogeneous diseases~\cite{SotosCMLhealthcare}, potentially non-invasively. 

However, despite significant advances, predictive deep learning models for medical image analysis are not immune to error, and severe consequences for the patient can occur if a clinician trusts erroneous predictions. A provided measure of uncertainty for each prediction is therefore essential to trust the model~\cite{EffectOfConfidence}. Although uncertainty is now commonly embedded in predictive medical image analysis (e.g. \cite{UncertaintySurvey,nair2020exploring,tousignant2019prediction}), it is not well-studied for precision medicine.  

\begin{figure}[t]
\centering
  \begin{subfigure}{0.47\textwidth}
            \includegraphics[clip,width=\textwidth]{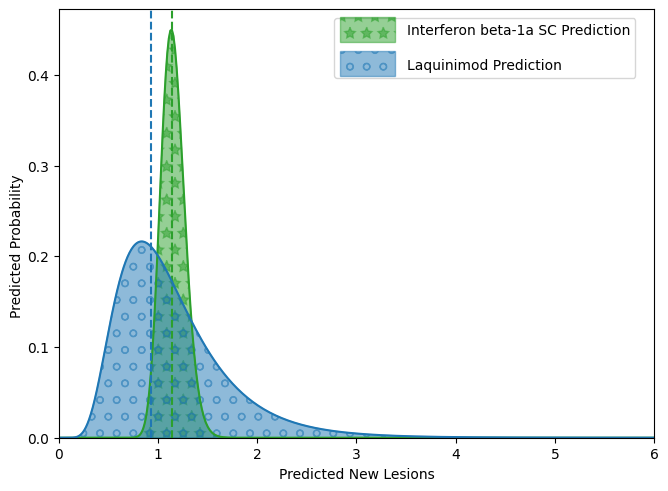}    
             \caption{} 
            \label{fig:PatientExamplesForOutcomesa}
  \end{subfigure}%
  \hspace*{\fill}   % maximize separation between the subfigures
  \begin{subfigure}{0.47\textwidth}
            \includegraphics[clip,width=\textwidth]{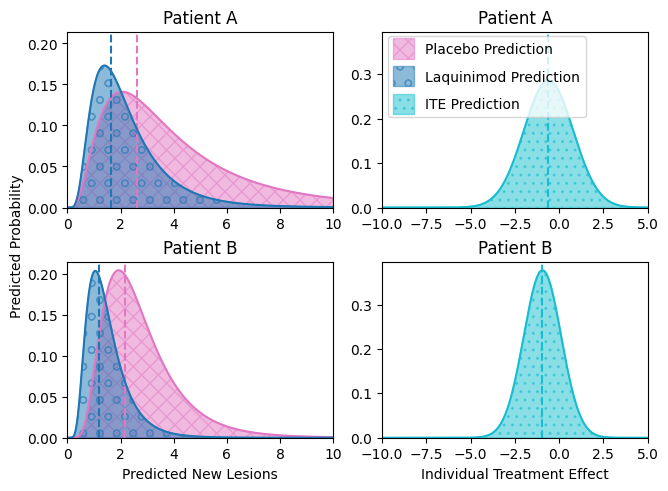}   
             \caption{} 
            \label{fig:PatientExamplesForOutcomesb}
  \end{subfigure}%
    \caption{(a) Probability distributions for a MS patient's predicted future new lesions on two different drugs  (\blue{laquinimod)} and \green{INFB-SC}). A patient might prefer \green{INFB-SC} if they are willing to make the tradeoff between slightly larger mean (dashed line) and lower variance (spread) in potential outcomes. (b) Predicted future outcomes for two patients for \blue{laquinimod} and \pink{placebo} drugs. Patients have similar expected (dashed line) \cyan{ITE} (difference between drugs), \blue{laquinimod}, and \pink{placebo} outcomes, but with different levels of confidence. Here, patient B is a better candidate for trial enrichment.}
    \label{fig:PatientExamplesForOutcomes}
\end{figure}

Image-based precision-medicine is highly relevant in multiple sclerosis (MS), a chronic disease characterized by the appearance over time of new or enlarging T2 lesions (NE-T2) on MRI  \cite{RRMSactivity,GadLesionsRelapses}. Several treatment options exist to suppress future NE-T2 lesions, but their level of efficacy and side effects are heterogeneous across the population \cite{MSheterogeneity}. Although one other model has been proposed for estimating the individual treatment effect (ITE) based on MR images \cite{JDFresponse}, it does not incorporate uncertainty. Fig.~\ref{fig:PatientExamplesForOutcomes} illustrates how knowledge of the model's uncertainty could improve treatment recommendations.

To integrate uncertainty into clinical decision making, new validation measures must be defined. The usual strategy for validating uncertainty estimates, discarding uncertain predictions \cite{AccuracyRejectionCurves,JessonCausalFailure} and examining performance on the remaining predictions, is not always appropriate when predicting treatment effects. For example, discarding uncertain predictions could result in discarding predictions for the most responsive of individuals. A better strategy for this individual would be to consider the level of response and uncertainty jointly when making a treatment decision. 

In this work, we present the first uncertainty-aware causal model for precision medicine based on medical images. We validate our model on a large, multi-center dataset of MR images from four different randomized clinical trials (RCTs) for MS. Specifically, we develop a multi-headed, Bayesian deep learning probabilistic model \cite{BayesModels} which regresses future lesion counts, a more challenging task than classification, but which provides more fine-grained estimates of treatment effect. We evaluate the model's uncertainty by showing correlation of predictive uncertainty on factual and counterfactual error, and demonstrate how to bound the treatment effect error using group-level ground truth data to evaluate its correlation with the predicted personalized treatment effect. We then show the use of incorporating predictive uncertainty to improve disease outcomes by better treatment recommendations. Lastly, we demonstrate how uncertainty can be used to enrich clinical trials and increase their statistical power \cite{TempleEnrichment}.

\section{Methods}
\subsection{Background on Individual Treatment Effect Estimation}
We frame precision medicine as a causal inference problem. Specifically, we wish to predict \textit{factual} outcomes (on the treatment a patient received), \textit{counterfactual} outcomes (on treatments a patient did not receive), as well as the individual treatment effect (ITE, the difference between the outcomes on two treatments). Let $X \in \mathbb{R}^d$ be the input features, $Y \in \mathbb{R}$ be the outcome of interest, and $T \in \{0, 1, ..., m\}$ be the treatment allocation with $t=0$ as a control (e.g. placebo) and the remaining are $m$ treatment options. Given a dataset containing triples $\mathcal{D} = \{(x^i, y^i, t^i)\}_{i=1}^n$, 
the ITE for patient $i$ and a drug $T=t$ can be defined using the Neyman/Rubin Potential Outcome Framework \cite{Rubin1974EstimatingCE} as $\text{ITE}_t=y_t - y_0$, where $y_t$ and $y_0$ represents \textit{potential} outcomes on treatment and control, respectively. The $\text{ITE}_t$ is an unobservable causal quantity because only one of the two potential outcomes is observed. The average treatment effect ($\textrm{ATE}_t$) is defined as $\mathbb{E}[\text{ITE}_t]=\mathbb{E}[y_t]-\mathbb{E}[y_0]$ and is an observable quantity. Treatment effect estimation in machine learning therefore relies on a related causal estimand, $\tau_t$:
\begin{equation}
    \tau_t(x) = \mathbb{E}[\text{ITE}_t|x] = \mathbb{E}[y_t-y_0|x] = \mathbb{E}[y_t|x] - \mathbb{E}[y_0|x].
\end{equation}
$\tau_t(x)$\footnote{Also known as conditional average treatment effect (CATE)} can be identified from RCT data (as in our case), where $(y_0, y_t) \independent T|X$ \cite{Gutierrez2017}. Individual treatment outcomes $y_t$ and $y_0$, and $\text{ITE}_t$, can therefore be estimated using machine learning models such that $\widehat{\text{ITE}}_t(x) = \hat{y}_t(x) - \hat{y}_0(x)$ \cite{MetalearnersforITE}.

%A Two-Model approach is frequently used to construct an estimator for $\tau_t(x)$, denoted $\hat{\tau}_t(x)$, and consists in training two models, $\hat{\mu_t}(x)$ and $\hat{\mu}_0(x)$, to estimate the conditional expectations $\mu_t$ and $\mu_0$, and subtracting their outputs such that $\hat{\tau}_t(x) = \hat{\mu_t}(x) - \hat{\mu}_0(x)$ (CITATION).

\subsection{Probabilistic Model of Individual Treatment Effects} 

In this work, we seek to learn the probability distribution of individual potential outcome predictions $\hat{y_t}(x)$ and the effect estimates $\widehat{\text{ITE}}_t(x)$. Let $\hat{y_t}(x) \sim \mathcal{N}(\hat{\mu_t}(x), \hat{\sigma_t}^2(x))$ be a normal distribution for potential outcome predictions whose parameters are outputs of a neural network. This probabilistic framework conveniently allows for propagating the uncertainty estimates for each potential outcome to an uncertainty estimate for personalized treatment effects. Assuming independence between the two Gaussian distributions, $\widehat{\text{ITE}}_t(x) \sim \mathcal{N}(\hat{\mu_t}(x)-\hat{\mu_{0}}(x), \hat{\sigma_{t}}(x)^2+\hat{\sigma_{0}}(x)^2)$. 

\begin{figure}[t]
    \centering
    \includegraphics[clip,width=\textwidth]{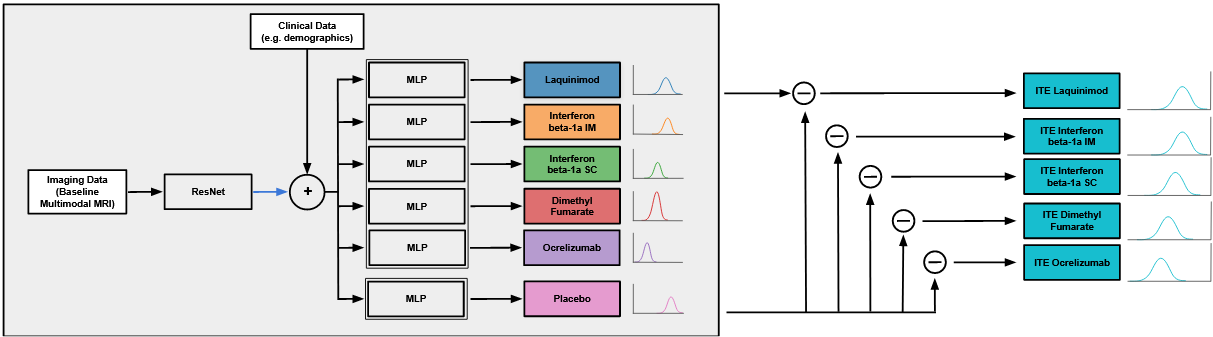}
    \caption{Multi-head ResNet architecture for treatment effect prediction (based on \cite{JDFresponse}). It is modified to generate probabilistic estimates of individual outcomes. Specific architecture details can be found in the Appendix.}
    \label{fig:ModelDiagram}
\end{figure}

For our specific context, the input $x$ to our model consists of multi-sequence patient MRI, lesion maps, and clinical and demographic features at baseline. The model is based on a multi-headed network for treatment response estimation \cite{JDFresponse,shalit2017TARNET}. Each head predicts $\hat{\mu_t}(x)$ and $\hat{\sigma_t}^2(x)$ for a particular treatment. For the case of MS, the model maximizes the log likelihood of the observed number of log NE-T2 lesions formed between 1 year and 2 years in the future (Fig. ~\ref{fig:ModelDiagram}).

\subsection{Evaluating Probabilistic Predictions}
\subsubsection{Bounds for the ITE Error}
\label{sec:Bounds} 
We can validate the quality of the estimated uncertainty for factual outcome predictions through the correlation between predictive uncertainty and Mean Squared Error (MSE) error. 
However, given that ground truth for the individual treatment effects are not available, we cannot compute MSE between $\text{ITE}_t$ and $\widehat{\text{ITE}}_t(x)$. 
% $\mathbb{E}[(\text{ITE}_t-\widehat{\text{ITE}}_t(x))^{2}]$. 
In this work, we choose to compute the upper and lower bounds for this MSE. We validate our uncertainty estimates by showing that selecting patients with the highest confidence in their predictions reduces the bounds on the ITE error. The bounds serve as an approximation to the true ITE error, and can validate models even if the ground truth ITE is not available.  We use the upper bound for the MSE as in \cite{shalit2017TARNET}. Jensen's inequality can be used to obtain a lower bound on the MSE as follows:
%\begin{equation}
%\begin{split}
%\mathbb{E}[(\tau_t (x)-\hat{\tau_t}(x))^{2}] \geq (\mathbb{E}[\tau_t (x)-\hat{\tau_t}.(x)])^{2}
% =(\mathbb{E}[\tau_t (x)]-\mathbb{E}[\hat{\tau_t}(x)])^{2}
%\end{split}
%\end{equation}

\begin{equation}
\begin{split}
\label{eq:bounds}
\mathbb{E}[(\text{ITE}_t -\widehat{\text{ITE}}_t(x))^{2}] \geq 
 (\mathbb{E}[\text{ITE}_t]-\mathbb{E}[\widehat{\text{ITE}}_t(x)])^{2}=(\textrm{ATE}_t-\mathbb{E}[\widehat{\text{ITE}}_t(x)])^{2}
\end{split}
\end{equation}
%(\mathbb{E}_x[(Y_t(x)-Y_0(x))-(\hat{Y_t}(x)-\hat{Y_0}(x))])^{2} & \\ 
%If we define a sub-group, $S_k$, of individuals with index $i\in \mathcal{S}_k$, then we can obtain an estimate of $\mathbb{E}[\tau_t (x)]$ by computing the average treatment effect ($\text{ATE}_t$)  for this sub-set: $\text{ATE}_t = \mathbb{E}[Y|T=t] - \mathbb{E}[Y|T=0]$. $\mathbb{E}[\hat{\tau_t}(x)]$ is calculated using the predicted ITE for sub-set $k$.

% and a measure of confounding variables

\subsubsection{Evaluating Individual Treatment Recommendations}
\label{sec:ITErecEval}
Predictive uncertainty can be used to improve treatment recommendations for the individual. Let $\pi(x^i,t^i) \in \{0,1\}$ be a treatment recommendation policy taking as input a patient's features $x_i$ and their factual treatment assignment $t^i$. The binary output of $\pi(x^i,t^i)$ denotes whether $t^i$ is recommended under $\pi$. In this work, we set $\pi$ to be a function of the model's predictions, $\hat{\mu_t}(x)$ and $\hat{\sigma_t}(x)$. For example, $\pi$ can be defined such that a treatment is recommended if the number of predicted NE-T2 lesions on a particular treatment are less than 2 \cite{MSguidelines}. An uncertainty aware policy could instead recommend a drug according to $P(\hat{y_t}(x)<2)$. The expected response under proposed treatments (ERUPT) \cite{ZhaoERUPT} can then be used to quantify the effectiveness of that policy:
 \begin{equation}
     \text{ERUPT}_{\pi}=\sum_{i=1}^n y^i*\pi(x^i,t^i)/\sum_{i=1}^n\pi(x^i,t^i)
     \label{eq:ERUPT}
 \end{equation}
 For the example of NE-T2 lesions, a lower value for ERUPT is better because there were fewer lesions on average for patients on the recommended treatment.

\subsubsection{Uncertainty for Clinical Trial Enrichment}
\label{sec:UncQuant}
Enriching a trial with predicted responders has been shown to increase statistical power in the context of MS \cite{FaletPMStreatmenteffect}. We measure the statistical power by the z-score: $\textrm{ATE}_t/\sqrt{\text{Var}(y_t)+\text{Var}(y_0)}$. 
% $\textrm{ATE}_t/\sqrt{\text{Var}_{i,t^i=t}(y^i)+\text{Var}_{i,t^i=0}(y^i)}$. 
Where $\text{Var}(y_t)$
% $\text{Var}_{i,t^i=t}(y^i)$
is the variance of factual outcomes on treatment $t$. The approach taken by \cite{FaletPMStreatmenteffect} achieves higher statistical power by selecting a subset of the population with larger $\textrm{ATE}_t$. Our proposed uncertainty-based enrichment selects patients with lower ITE uncertainty ($\hat{\sigma_t}(x)^2+\hat{\sigma_0}(x)^2$), with the goal of reducing the population variance ($\text{Var}(y_t)+\text{Var}(y_0)$). The benefit of this approach is most apparent if we inspect a specific population (defined by a particular value for ATE in the numerator).

\section{Experiments and Results}
\subsection{Dataset}
The dataset is composed of patients from four randomized clinical trials: BRAVO \cite{BRAVO}, OPERA 1 \cite{OPERA}, OPERA 2 \cite{OPERA}, and DEFINE \cite{DEFINE}. Each trial enrolled patients with relapsing-remitting MS. Each patient sample consists of multi-sequence patient MRI (T1 weighted pre-contrast, T1 weighted post-contrast, FLAIR, T2-weighted, and proton density weighted), lesion maps (T2 hyperintense and gadolinium-enhancing lesions), as well as relevant clinical and demographic features (age, sex, expanded disability status scale scores \cite{Kurtzke1444EDSS}) at baseline. The number of NE-T2 lesions between 1 and 2 years after trial initiation  were provided for each patient.  Excluding patients with incomplete data resulted in a dataset with $n=2389$ patients. In total the dataset contains the following treatment arms: \pink{placebo} ($n=406$), \textcolor[HTML]{1f77b4}{laquinimod} ($n=273$) , \textcolor[HTML]{ff7f0e}{interferon beta-1a intramuscular (INFB-IM)}  ($n=304$), \textcolor[HTML]{2ca02c}{interferon beta-1a subcutaneous (INFB-SC)} ($n=564$), \textcolor[HTML]{d62728}{dimethyl fumarate (DMF)}  ($n=225$), and \textcolor[HTML]{9467bd}{ocrelizumab} ($n=627$). We perform 4 fold nested cross validation on this dataset. \cite{VabalasNested}

\subsection{Evaluation of Factual Predictions and Uncertainty Estimation}
\label{sec:FacCfacUncCal}
\begin{figure}[t]
  \begin{subfigure}{0.33\textwidth}
            \includegraphics[clip,width=\textwidth]{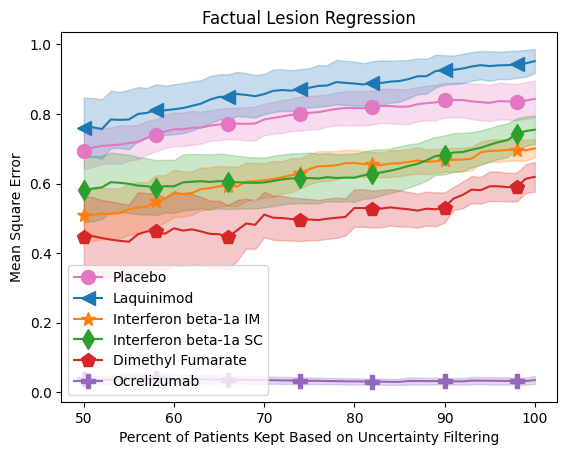}    \caption{} \label{fig:MSEcontinuousErrora}
  \end{subfigure}%
  \hspace*{\fill}   % maximize separation between the subfigures
  \begin{subfigure}{0.33\textwidth}
            \includegraphics[clip,width=\textwidth]{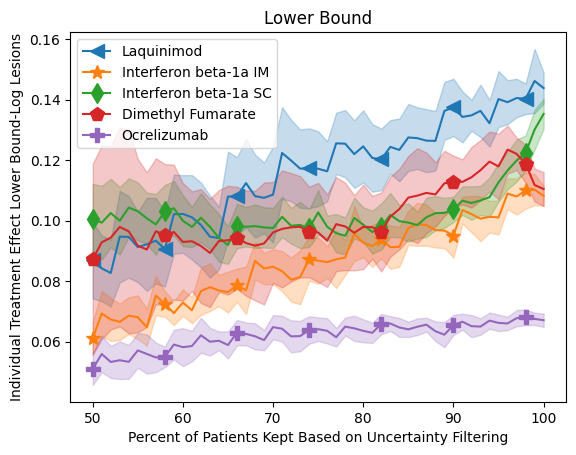}    \caption{} \label{fig:MSEcontinuousErrorb}
  \end{subfigure}%
  \hspace*{\fill}   % maximizeseparation between the subfigures
  \begin{subfigure}{0.33\textwidth}
            \includegraphics[clip,width=\textwidth]{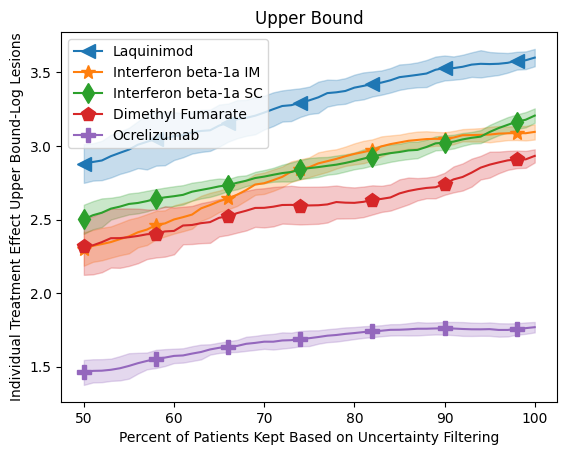} \caption{} \label{fig:MSEcontinuousErrorc}
  \end{subfigure}
    \caption{(a)  MSE for the log-lesion outcome as a function of predictive uncertainty. MSE is plotted separately for each treatment, using only patients who factually received the particular treatment. Uncertainty is computed according to the variance of the normal distribution predicted by the model, and the $x$-axis refers to the percent of kept patients based on uncertainty filtering (i.e. At 100, all patients are kept when computing MSE). (b) Lower bound for the ITE error as a function of predicted ITE uncertainty. (c) Upper bound for the ITE error as a function of predicted ITE uncertainty.}
        \label{fig:MSEcontinuousError}

    \end{figure}

Each patient is given a single treatment. The MSE for the future log-NE-T2 lesion count on the observed (factual) treatment and $\hat{\mu}_t(x)$ is used as a measure of the model's predictive accuracy. Taking all treatments in aggregate, the model achieves an MSE of $0.59 \pm 0.03$. Separating each treatment, it achieves an error of $0.84 \pm 0.10$ for \pink{placebo}, $0.95 \pm 0.07$ for \blue{laquinimod},  $0.70 \pm 0.05$ for \orange{INFB-IM},  $0.76 \pm 0.08$ for \green{INFB-SC}, $0.62 \pm 0.08$ for \red{DMF}, and $0.04 \pm 0.02$ for \purple{ocrelizumab}. Next we evaluate the correlation between the model error and the predicted variance. An accurate uncertainty estimate should be positively correlated with prediction accuracy \cite{AccuracyRejectionCurves}.  This relationship is shown in  Fig.~\ref{fig:MSEcontinuousErrora}, where the MSE for the factual predictions decreases as we select a sub-group of patients with lower predictive uncertainty.

Next, we examine the results for the ITE error. Fig.~\ref{fig:MSEcontinuousErrorb} and Fig.~\ref{fig:MSEcontinuousErrorc} show the upper and lower bounds (Eq.~\ref{eq:bounds}). Similarly to the factual error, the lower bound and upper bound on the ITE error decrease with decreasing ITE uncertainty.
% Because we lack ground-truth ITEs, the correlation for the ITE-uncertainty cannot be calculated in the same way. Instead, a lower and upper bound for the ITE error (see Sec. \ref{sec:Bounds}) are shown as a function of ITE uncertainty (estimated from the model as $\hat{\sigma_{t}}(x)^2+\hat{\sigma_{0}}(x)^2$). 

\subsection{Uncertainty for Individual Treatment Recommendations}
\label{sec:ClinDecUnc}
\begin{figure}[t]
  \begin{subfigure}{0.47\textwidth}
            \includegraphics[clip,width=\textwidth]{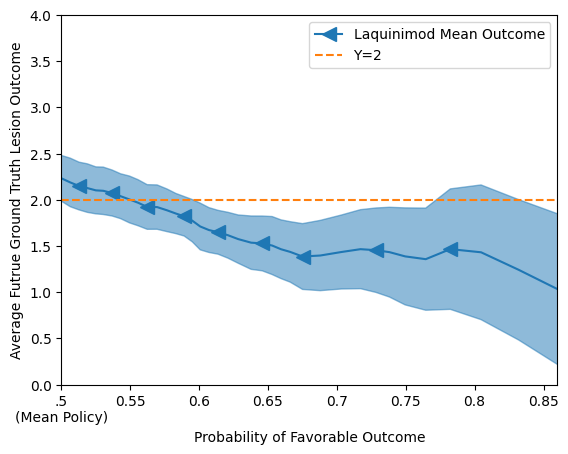}    
            \caption{} 
            \label{fig:LaqPredictiveUncertaintya}
  \end{subfigure}%
  \hspace*{\fill}   % maximize separation between the subfigures
  \begin{subfigure}{0.47\textwidth}
            \includegraphics[clip,width=\textwidth]{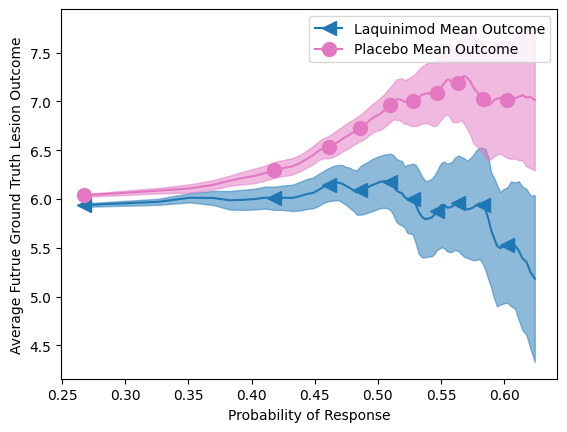}   
            \caption{} 
            \label{fig:LaqPredictiveUncertaintyb}
  \end{subfigure}%
    \caption{Average factual future NE-T2 lesion count for patients recommended \blue{laquinimod} under different uncertainty-aware policies. (a) The policy recommends \blue{laquinimod} based on the probability that \blue{laquinimod} will lead to fewer than 2 NE-T2 lesions in the future. 
    (b) The policy recommends \blue{laquinimod} based on the probability of response (defined as having fewer lesions on \blue{laquinimod} than on \pink{placebo}).}
    \label{fig:LaqPredictiveUncertainty}
\end{figure}

The effect of integrating uncertainty into treatment recommendations is evaluated by defining a policy using this uncertainty (Eq. \ref{eq:ERUPT}). Here, we report outcomes on the lesion values (as opposed to log-lesions) for interpretability. In Fig.~\ref{fig:LaqPredictiveUncertaintya}, a treatment, \blue{laquinimod}\footnote{Results on other treatments can be found in the Appendix.}, is recommended if the predicted probability of having fewer than 2 NE-T2 lesions is greater than a threshold $k$: $P(\hat{y_t}(x)<2)>k$. A policy requiring greater confidence indeed selects patients who more often have fewer than 2 lesions. It is worth noting that \blue{laquinimod} was not found to be efficacious at the whole group level in clinical trials \cite{BRAVO} and is therefore not approved, but this analysis shows that using personalized recommendations based on uncertainty can identify a sub-group of individuals that can benefit.

In Fig.~\ref{fig:LaqPredictiveUncertaintyb}, a treatment effect-based policy is used such that \blue{laquinimod} is recommended if the probability of any treatment response  is greater than a threshold $k$: $P(\widehat{\text{ITE}}_t(x)\leq0)>k$. As certainty in response grows, the difference between the treated and \pink{placebo} groups grows suggesting an uncertainty aware policy better identifies patients for which the drug will have an effect. 

\begin{figure}[hbt]
  \begin{subfigure}{0.47\textwidth}
            \includegraphics[clip,width=\textwidth]{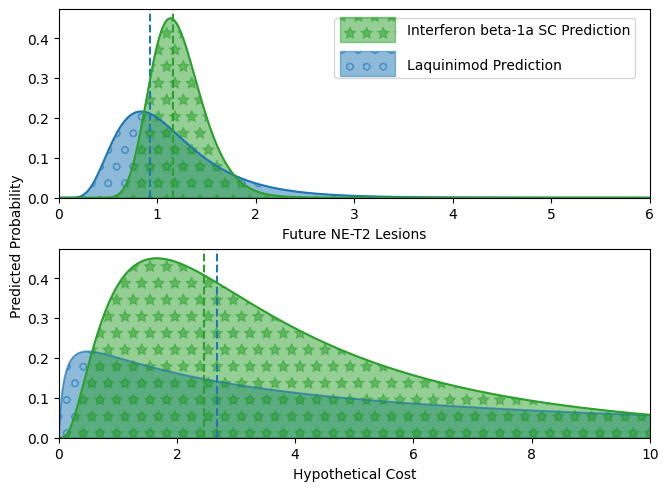}    
            \caption{} 
            \label{fig:Costexamplea}
  \end{subfigure}%
  \hspace*{\fill}   % maximize separation between the subfigures
  \begin{subfigure}{0.47\textwidth}
            \includegraphics[clip,width=\textwidth]{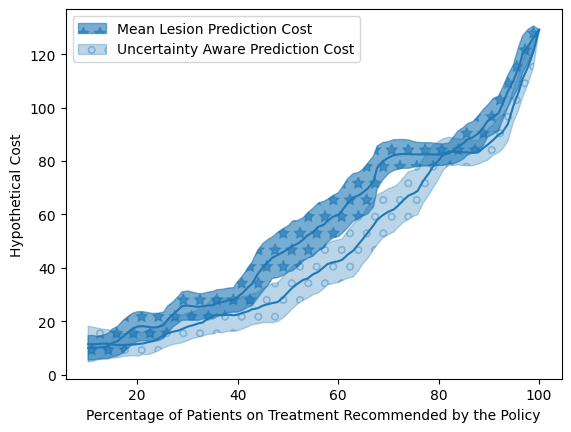}   
            \caption{} 
            \label{fig:Costexampleb}
  \end{subfigure}%
    \caption{(a) Example of the predicted outcomes for a single patient on two drugs (\blue{laquinimod} and \green{INFB-SC}) before [top] and after [bottom] a hypothetical cost transformation. Note that the transformation causes the recommended treatment (as defined by the mean of the distribution, see dashed line) to switch from \blue{laquinimod} to \green{INFB-SC}. (b) The expected cost under the mean and uncertainty-aware policies at the level of the entire \blue{laquinimod} cohort.}
    \label{fig:QualExamplecost99}
\end{figure}

Uncertainty can be useful when we wish to attribute a cost, or risk, to a certain range of outcomes. In our case, we assume a hypothetical non-linear cost $c$ for having more NE-T2 lesions, where $c = (\text{NE-T2 lesions}+1)^{2}$. Fig.~\ref{fig:Costexamplea} describes a case where the recommended treatment (in terms of the mean) changes if this cost transformation is applied. In this case, the shape of the distribution over possible outcomes (which informs our uncertainty about this outcome) affects how much the mean of the distribution shifts under this transformation. This analysis is extended to the entire \blue{laquinimod} cohort in Fig.~\ref{fig:Costexampleb}. We compute the average cost (Eq. \ref{eq:ERUPT}) rather than the number of future NE-T2 lesions (as in Fig.~\ref{fig:LaqPredictiveUncertaintya}) for two types of policies. In the uncertainty-aware policy, the predicted distribution is used to make the treatment decision, whereas for the mean policy, the decision is based on only on $\hat{\mu_t}(x)$. As expected, uncertainty-aware recommendations incur a lower expected cost across the entire cohort compared to the mean policy. The advantage is most visible for intermediate values on the $x$-axis, because at the far right all patients are recommended \blue{laquinimod}, and at the far left patients have closer to 0 NE-T2 lesions on average and the magnitude of the improvement due to the uncertainty-aware policy lessens.

\label{sec:predict_uncert}

\subsection{Uncertainty for Clinical Trial Enrichment}
\begin{figure}[h!]
    \centering
        \includegraphics[clip,width=.4\textwidth]{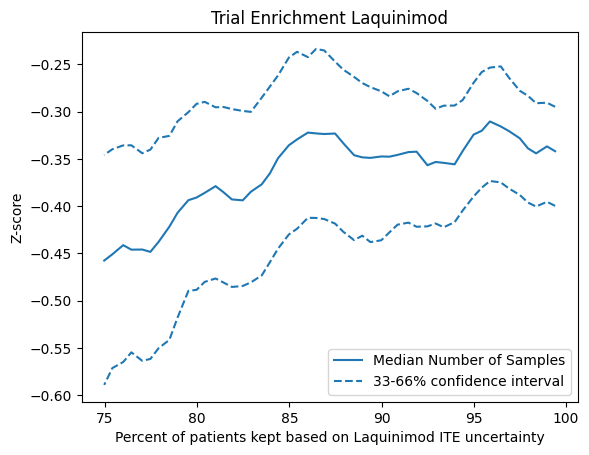}   
\caption{To isolate the effect of uncertainty on enrichment, we fixed the ATE to be equal to 0 to -2 NE-T2 lesions by including patients with fixed placebo ($2<\hat{y_0}(x)<3$) and treatment ($1<\hat{y_t}(x)<2$) outcomes. The z-score then decreases for patient groups with smaller predicted ITE uncertainty.}
  \label{fig:ITEtrialEnrichment}
\end{figure}

Uncertainty estimation can also be useful when selecting a sub-population of to enroll in a clinical trial, in a technique called predictive enrichment \cite{EfficienyOfTargetDesignsforRCTs}. Fig. \ref{fig:PatientExamplesForOutcomesb}, shows an example where two patients have similar estimated future lesions but different ITE uncertainties. For trial enrichment, the second patient is more likely to experience a significant effect from this drug, and therefore enriching the trial with such patients could increase it's statistical power to detect an effect if done appropriately (see Sec. \ref{sec:UncQuant}). In Fig. \ref{fig:ITEtrialEnrichment} we show the effect of uncertainty-aware trial enrichment. For a population with a particular effect size, we remove patients (right to left) with high ITE uncertainty and compute the z-score between the untreated and treated populations for the remaining groups. As expected, groups with smaller average ITE uncertainty have greater statistical differences (lower z scores).

\section{Conclusion}
    In this work, we present a novel, causal, probabilistic, deep learning framework for image-based precision medicine. Our multi-headed architecture produces distributions over potential outcomes on multiple treatment options and a distribution over personalized treatment effects. We evaluate our model on a real-world, multi-trial MS dataset, where we demonstrate quantitatively that integrating the uncertainties associated with each prediction can improve treatment-related outcomes in several real clinical scenarios compared to a simple mean prediction. The evaluation methods used in this work are agnostic to the method of uncertainty quantification which permits flexibility in the choice of measure. Overall, this work has the potential to greatly increase trust in the predictions of causal models for image-based precision medicine in the clinic.

\section*{Acknowledgement} This investigation was supported by the International Progressive Multiple Sclerosis Alliance (PA-1412-02420), the companies who generously provided the data: Biogen, BioMS, MedDay, Novartis, Roche / Genentech, and Teva, the Canada Institute for Advanced Research (CIFAR) AI Chairs program, the Natural Sciences and Engineering Research Council of Canada, the MS Society of Canada, Calcul Quebec, and the Digital Research Alliance of Canada. The authors would like to thank L. Collins and M. Dadar for preprocessing the MRI data, Z. Caramanos, A. M. Pinzon, C. Guttmann and I, Mórocz for collating the clinical data, and S. Narayanan and M.-P. Sormani for their MS expertise.

% Louis Collins and Mahsa Dadar for preprocessing the MRI data, Zografos Caramanos, Alfredo Morales Pinzon, Charles Guttmann and István Mórocz for collating the clinical data, and Sridar Narayanan. Maria-Pia Sormani for their MS expertise.

 %\newpage
\bibliographystyle{splncs04}
\bibliography{bibliography}

\begin{thebibliography}{10}
\providecommand{\url}[1]{\texttt{#1}}
\providecommand{\urlprefix}{URL }
\providecommand{\doi}[1]{https://doi.org/#1}

\bibitem{UncertaintySurvey}
Abdar, M., et~al.: A review of uncertainty quantification in deep learning:
  Techniques, applications and challenges. Information Fusion  \textbf{76},
  243--297 (2021)

\bibitem{JDFresponse}
Durso-Finley, J., et~al.: Personalized prediction of future lesion activity and
  treatment effect in multiple sclerosis from baseline mri (2022)

\bibitem{FaletPMStreatmenteffect}
Falet, J.P.R., et~al.: Estimating treatment effect for individuals with
  progressive multiple sclerosis using deep learning  (2021)

\bibitem{MSguidelines}
Freedman, M., et~al.: Treatment optimization in multiple sclerosis: Canadian ms
  working group recommendations. Canadian Journal of Neurological Sciences /
  Journal Canadien des Sciences Neurologiques  \textbf{47},  1--76 (04 2020)

\bibitem{Gutierrez2017}
Gutierrez, P., et~al.: Causal inference and uplift modelling: A review of the
  literature. vol.~67, pp. 1--13. PMLR (12)

\bibitem{OPERA}
Hauser, S.L., et~al.: Ocrelizumab versus interferon beta-1a in relapsing
  multiple sclerosis. New England Journal of Medicine  \textbf{376}(3),
  221--234 (2017)

\bibitem{DEFINE}
Havrdova, E., et~al.: {{O}ral {B}{G}-12 (dimethyl fumarate) for
  relapsing-remitting multiple sclerosis: a review of {D}{E}{F}{I}{N}{E} and
  {C}{O}{N}{F}{I}{R}{M}. {E}valuation of: {G}old {R}, {K}appos {L}, {A}rnold
  {D}, and others {P}lacebo-controlled phase 3 study of oral {B}{G}-12 for
  relapsing multiple sclerosis.} Expert Opin Pharmacother  \textbf{14}(15),
  2145--2156 (Oct 2013)

\bibitem{JessonCausalFailure}
Jesson, A., et~al.: Identifying causal effect inference failure with
  uncertainty-aware models  (2020)

\bibitem{GadLesionsRelapses}
Kappos, L., et~al.: Predictive value of gadolinium-enhanced magnetic resonance
  imaging for relapse rate and changes in disability or impairment in multiple
  sclerosis: a meta-analysis. Lancet (London, England)  \textbf{353},  964--969
  (1999)

\bibitem{Kurtzke1444EDSS}
Kurtzke, J.F.: Rating neurologic impairment in multiple sclerosis. Neurology
  \textbf{33} (1983)

\bibitem{MetalearnersforITE}
Künzel, S.R., Sekhon, J.S., Bickel, P.J., Yu, B.: Metalearners for estimating
  heterogeneous treatment effects using machine learning. Proceedings of the
  National Academy of Sciences  \textbf{116}(10),  4156--4165 (feb 2019)

\bibitem{MSheterogeneity}
Lucchinetti, C., et~al.: Heterogeneity of multiple sclerosis lesions:
  Implications for the pathogenesis of demyelination. Annals of neurology
  \textbf{47},  707--17 (07 2000)

\bibitem{BayesModels}
MacKay, D.J.C.: {A Practical Bayesian Framework for Backpropagation Networks}.
  Neural Computation  \textbf{4}(3),  448--472 (05 1992)

\bibitem{AccuracyRejectionCurves}
Nadeem, M.S.A., et~al.: Accuracy-rejection curves (arcs) for comparing
  classification methods with a reject option. In: Proceedings of the third
  International Workshop on Machine Learning in Systems Biology. Proceedings of
  Machine Learning Research, vol.~8, pp. 65--81. PMLR, Ljubljana, Slovenia
  (05--06 Sep 2009)

\bibitem{nair2020exploring}
Nair, T., et~al.: Exploring uncertainty measures in deep networks for multiple
  sclerosis lesion detection and segmentation. Medical image analysis
  \textbf{59},  101557 (2020)

\bibitem{Rubin1974EstimatingCE}
Rubin, D.B.: Estimating causal effects of treatments in randomized and
  nonrandomized studies. Journal of Educational Psychology  \textbf{66},
  688--701 (1974)

\bibitem{RRMSactivity}
Rudick, R., et~al.: Significance of t2 lesions in multiple sclerosis: A 13-year
  longitudinal study. Annals of neurology  \textbf{60},  236--42 (08 2006)

\bibitem{SotosCMLhealthcare}
Sanchez, P., et~al.: Causal machine learning for healthcare and precision
  medicine. Royal Society Open Science  \textbf{9}(8) (2022)

\bibitem{shalit2017TARNET}
Shalit, U., et~al.: Estimating individual treatment effect: generalization
  bounds and algorithms (2017)

\bibitem{EfficienyOfTargetDesignsforRCTs}
Simon, R., Maitournam, A.: Evaluating the efficiency of targeted designs for
  randomized clinical trials. Clinical cancer research : an official journal of
  the American Association for Cancer Research  \textbf{10},  6759--63 (11
  2004)

\bibitem{TempleEnrichment}
Temple, R.: Enrichment of clinical study populations. Clinical Pharmacology \&
  Therapeutics  \textbf{88}(6),  774--778 (2010)

\bibitem{tousignant2019prediction}
Tousignant, A., et~al.: Prediction of disease progression in multiple sclerosis
  patients using deep learning analysis of mri data. In: International
  conference on medical imaging with deep learning. pp. 483--492. PMLR (2019)

\bibitem{VabalasNested}
{Vabalas}, A., et~al.: {Machine learning algorithm validation with a limited
  sample size}. PLoS ONE  \textbf{14}(11) (Nov 2019)

\bibitem{RueckertCausalReview}
Vlontzos, A., et~al.: A review of causality for learning algorithms in medical
  image analysis  (2022)

\bibitem{BRAVO}
Vollmer, T.L., et~al.: {{A} randomized placebo-controlled phase {I}{I}{I} trial
  of oral laquinimod for multiple sclerosis}. J Neurol  \textbf{261}(4),
  773--783 (Apr 2014)

\bibitem{EffectOfConfidence}
Zhang, Y., et~al.: Effect of confidence and explanation on accuracy and trust
  calibration in ai-assisted decision making. CoRR  (2020)

\bibitem{ZhaoERUPT}
Zhao, Y., et~al.: Uplift modeling with multiple treatments and general response
  types (2017)

\end{thebibliography}
\end{document}